\documentclass{article}

\usepackage[preprint]{neurips_2021_arxiv}

\usepackage[utf8]{inputenc} %
\usepackage[T1]{fontenc}    %
\usepackage{hyperref}       %
\usepackage{url}            %
\usepackage{booktabs}       %
\usepackage{amsfonts}       %
\usepackage{nicefrac}       %
\usepackage{microtype}      %
\usepackage{xcolor}         %

\usepackage[american]{babel}

\usepackage{natbib} %
\usepackage{mathtools} %
\usepackage{booktabs} %
\usepackage{tikz} %

\usepackage{relsize}

\usepackage[utf8]{inputenc}

\usepackage{amsmath, amsthm}
\usepackage{mathtools}
\usepackage{amsfonts,amssymb,dsfont,wrapfig}
\usepackage{xfrac}
\usepackage{tikz, tikz-network}
\usepackage{subcaption} 
\usepackage[rightcaption]{sidecap}
\usepackage{pifont}%
\DeclareMathAlphabet{\mathcal}{OMS}{cmsy}{m}{n}

\tikzstyle{dot}=[circle,fill,inner sep=2.5pt]  %
\tikzstyle{dgraph}=[->, line width=1.5pt]
\usetikzlibrary{bayesnet}

\usepackage{todonotes}
\usepackage{bbm,cancel}
\newtheorem{theorem}{Theorem}

\renewcommand{\eqref}[1]{Eq. (\ref{#1})}
\newcommand{\figref}[1]{Fig. \ref{#1}}
\newcommand{\secref}[1]{Sect. \ref{#1}}

\newcommand{\E}[2]{\mathop{\mathlarger{\mathbb{E}} }_{#1} #2}

\newcommand{\xli}{{x_{<i}}}

\newcommand{\sign}{{sgn}}

\newcommand{\DD}{{\cal D}}
\newcommand{\NN}{{\cal N}}
\newcommand{\OO}{{\cal O}}
\newcommand{\GG}{{\cal G}}
\newcommand{\MM}{{\cal M}}
\newcommand{\XX}{{X}}
\newcommand{\ZZ}{{\mathcal Z}}

\newcommand{\II}{{\cal I}}

\newcommand{\indep}{\perp \!\!\! \perp}
\newcommand{\pa}{{\text{pa}}}
\newcommand{\cpd}{CPD}
\newcommand{\cpds}{CPDs}
\newcommand{\prm}{\theta}
\newcommand{\PI}{\text{preq}}

\newcommand{\cmark}{\ding{51}}%
\newcommand{\xmark}{\ding{55}}%

\newcommand{\vcentered}[1]{
\begin{tikzpicture}
\clip (-.15,-.4) rectangle (.15,.4);
\node at (0,0) {#1};
\end{tikzpicture}}

\newcommand{\wvcentered}[1]{
\begin{tikzpicture}
\clip (-.4,-.4) rectangle (.4,.4);
\node at (0,0) {#1};
\end{tikzpicture}}

\newcommand{\method}{prequential scoring}
\newcommand{\Method}{Prequential scoring}

\title{Prequential MDL for Causal Structure Learning \\ with Neural Networks}

\author{%
    Jörg Bornschein \\
    \And
    Silvia Chiappa \\
    \And
    Alan Malek \\
    \And
    Rosemary Nan Ke \\ 
    \hskip-9.5cm DeepMind, London \\
    \hskip-9cm\texttt{\{bornschein,csilvia,alanmalek,nke\}@deepmind.com}
}

\begin{document}

\maketitle

\begin{abstract}
Learning the structure of Bayesian networks and causal relationships from observations is a common goal in several areas 
of science and technology. 
We show that the prequential minimum description length principle (MDL) can be used to derive a practical scoring function 
for Bayesian networks when flexible and overparametrized neural networks are used to model the conditional probability 
distributions between observed variables. 
MDL represents an embodiment of Occam's Razor and we obtain plausible and parsimonious graph structures 
without relying on sparsity inducing priors or other regularizers which must be tuned.
Empirically we demonstrate competitive results on synthetic and real-world data.  
The score often recovers the correct structure even in the presence of strongly nonlinear 
relationships between variables; a scenario were prior approaches struggle and usually fail. 
Furthermore we discuss how the the prequential score relates to 
recent work that infers causal structure from the speed of adaptation
when the observations come from a source undergoing distributional 
shift.

\end{abstract}

\section{Introduction}
Bayesian networks are a powerful probabilistic framework based on a graphical representation of statistical relationships between random variables. Inferring the Bayesian network structure that best represents a dataset not only allows to use the network to perform probabilistic, and possibly causal, reasoning but can also provide substantial illumination about the domain under consideration. This paper considers the problem of structure learning in settings in which modern, possibly overparametrized, neural networks are used to model the Bayesian network conditional distributions. 

Recent effort on structure learning with modern neural networks has focused on improving scalability w.r.t. the number of variables by relaxing the discrete search problem over structures to a continuous optimization problem \citep{zheng2018dags,yu2019dag,zheng2020learning}. Whilst enabling the use of large structures, the regularized maximum-likelihood score used to rank structures makes these methods prone to overfitting random fluctuations and sensitive to the regularizer. 

We propose an approach to ranking structures based on the minimum description length (MDL) principle. Motivated by fundamental ideas in data-compression, information theory, as well as philosophical notions like Occam's razor, the MDL principle posits that models which lead to compact and parsimonious descriptions of the data are more plausible. In the context of structure learning, this criterion induces a preference for more compact and simpler structures as more plausible explanations of the data generation mechanism. Many traditional scores, such as AIC \citep{akaike1973information}, BIC \citep{schwarz1978estimating}, marginal likelihood \citep{heckerman1995learning}, and more recent scores \citep{silander2018quotient}, can be seen as implementations of the MDL principle. However, some of these scores are approximations that, especially when applied to overparametrized neural networks, can lead to poor empirical performance \citep{silander2008factorized}. In addition, many such scores can only be applied to simple model families, and therefore might not be suitable to modelling complex nonlinear relationships in the data.

We propose using the prequential plug-in score, which evaluates conditional distributions by their sequential predictive performance, and gives an approach to ranking structures that balances fit to the data with overfitting without the need for an explicit sparsity inducing priors or regularizer. We provide a specific method for implementing the score with modern neural networks. We demonstrate on both artificial and real-world data that our method often retrieves the  data-generating structure and is robust to neural network hyperparameter selection.

\section{Structure Learning with Prequential MDL}
\label{sec:UD}
Our approach to learning the structure of a Bayesian network is to rank structures by measuring the complexity of the associated \emph{conditional probability distributions} (\cpds) through the \emph{minimum description length} (MDL) principle \citep{grunwald2004tutorial,grunwald2007minimum}.  
In particular, we propose the use of the prequential plug-in score and an implementation of this score for the case in which modern neural networks are used to model the \cpds. Before describing our method in detail, we give an introduction into Bayesian networks and structure learning with MDL.

\paragraph{Bayesian Networks (BNs).}

\begin{wrapfigure}[4]{r}{0.29\textwidth}
\vspace{-0.6cm}
\scalebox{0.9}{
\begin{tikzpicture}[dgraph]
\hspace{-0.3cm}
\node[] (G) [label= north:$(\GG\text{, }p)$] at (-0.95,0.1) {};
\node[dot] (A) [label= north:$X^1$, label= south:$p(X^1)$] at (-1.8,0) {};
\node[dot] (C) [label= north:$X^2$, label= south:$p(X^2| X^1 \text{, } X^3)$] at (0,0) {};
\node[dot] (B) [label= north:$X^3$, label= south:$p(X^3)$] at (1.8,0) {};
\draw[line width=0.6pt](A)--(C);
\draw[line width=0.6pt](B)--(C);
\end{tikzpicture}}
\end{wrapfigure}

A \emph{Bayesian network} \citep{pearl1988probabilistic,pearl2000causality,cowell2001probabilistic,kollerl2009probabilistic} is a \emph{directed acyclic graph} (DAG) $\GG$ whose nodes $X^1,\ldots, X^D$ represent random variables and links express statistical dependencies among them. Node $X^d$ is associated with \cpd~$p(X^d \,|\, \pa(X^d))$, where $\pa(X^d)$ denote the \emph{parents} of $X^d$, namely the nodes with a link into $X^d$. The joint distribution of all nodes is given by the product of all \cpds, i.e. $p(X^1, \dots, X^D \,|\,\GG) = \prod_{d=1}^D p(X^d \,|\, \pa(X^d))$. We make the common assumption that each \cpd~is parametrized by separate parameters. 
The set of BNs that encode the same set of conditional independence assumptions forms a \emph{Markov equivalence class}. 
A BN can be given a causal semantic by interpreting a link between two nodes as expressing causal rather than statistical dependence.

\paragraph{Score-based Structure Learning with MDL.} 
Let $p^*$ be a joint distribution over $D$ random variables with joint domain $\mathcal X$, and 
let $\DD=\{x_i:= (x_i^1,\ldots x_i^D)\}_{i=1}^n$ be a dataset of $n$ i.i.d samples from $p^*$. The goal of \emph{structure learning} is to infer the DAG $\GG$, referred to as \emph{structure}, or the Markov equivalence class that best represents $\DD$. 

We focus on score-based approaches that rank structures w.r.t. some scoring metric \citep{heckerman1999tutorial,drton2017structure}. A na\"ive score is the \emph{maximum log-likelihood} $\max_{\prm\in\Theta_\GG} \,\log p(\DD\,|\,\prm,\GG)$, which ignores model complexity and results in a preference for dense and complex structures that do not generalize well. A simple approach to account for model complexity is to add to the log-likelihood a regularization term that can depend on the dimension of the parameters $\dim(\prm)$ and on the size of the dataset $n$---the two most common penalty terms are $-\dim(\prm)$ (AIC) and $-0.5\log(n)\dim(\prm)$ (BIC). A more sophisticated approach is to instead integrate out $\theta$, which gives the \emph{log-marginal likelihood} $\log \int_{\Theta_\GG} p(\DD\,|\, \prm, \GG) p(\prm\,|\,\GG) d\prm$. 
Both these approaches can be described within the unifying framework of MDL.

The MDL framework is based on the principle that the model that yields the shortest description of the data is also the most plausible. In the context of structure learning, the MDL principle prescribes that we pick the model class ${\MM}_{\GG}=\{p(\cdot|\theta,\GG):\theta\in \Theta_{\GG}\}$ from the set $\{\MM_\GG:\GG\in G\}$ which leads to the most compact representation of the dataset $\DD$ with a code derived from ${\MM}_{\GG}$. 
Considering the close relationship between code-lengths and probability distributions this means selecting the model class under which 
the data has the highest likelihood. 
From this perspective maximum-likelihood $\log p(\DD|\hat\theta^{\text{MLE}}(\DD),\GG)$, where $\hat\theta^{\text{MLE}}(\DD) := \arg\max_\theta \log\,p(\DD|\theta,\GG)$ alone cannot be the basis for a code because it does not normalize to one, it is thus not a distribution over $\DD$ which 
precludes the existence of a code with the corresponding code-length. 
Maximum likelihood can only be used as a basis for a code if $\theta^{\text{MLE}}$ is known a-priori.

Instead, the MDL literature suggests to use the log-likelihood of the \emph{universal distribution} $\overline p(\cdot|\GG)$ of $\MM_\GG$ on $\mathcal X^n$, defined as
\begin{align}
\overline p(\cdot|\GG):=\arg\max_{q}\min_{\ZZ \in \mathcal X^n}\left(\log q(\ZZ)-\log p(\ZZ|\hat\theta^{\text{MLE}}(\ZZ),\GG)\right)
\label{eq:universal-dist}
\end{align}
to quantify the minimal code-length under a model class $\MM_\GG$. The MDL structure selection rule is 
\begin{equation}
\GG_{\text{MDL}}(\DD) := \arg\max_{\GG\in G} \, \log \overline p(\DD\,|\, \GG)
\label{eq:MDLselectionRule}
\end{equation}
because the model class with the most compact representation corresponds to the model class with highest 
$\overline p(\DD|\GG)$ (see \cite{grunwald2019minimum}).
The universal distribution is a probability distribution that, in some sense, summarizes how well $\MM_\GG$ fits data: it places large probability on $\DD$ only if there is a distribution $p(\cdot|\theta,\GG)\in\MM_\GG$ that places large probability on $\DD$. 
The requirement that $\overline p(\cdot|\GG)$ must normalize to one naturally induces complexity regularization. For a model class that is very expressive (e.g. $\GG$ is fully connected)  $\log\, p(\ZZ|\hat\theta^{\text{MLE}}(\ZZ),\GG)$ is large for many values of $\ZZ$, and therefore the universal distribution must spread its mass across much of $\mathcal X^n$. This implies that $\log \overline p(\DD|\GG)$ for the observed dataset $\DD$ cannot be high.
On the other hand, for a model class that is not as expressive (e.g, $\GG$ includes only few links), $\log p(\ZZ|\hat\theta^{\text{MLE}}(\ZZ),\GG)$ is large only for data that are compatible with its graph structure and the universal distribution can have much higher log-likelihood on such data.
Using the universal distribution for structure selection leads to favoring structures that have expressiveness for data similar to $\DD$ but do not waste expressiveness on dissimilar data. 

\subsection{Prequential Plug-in Score}
\label{sec:PPI}
Equation \ref{eq:universal-dist} provides a prescriptive definition of the universal distribution required by
\eqref{eq:MDLselectionRule} to compute the score $\log \overline p(\DD\,|\, \GG)$.
Several constructive definitions have been proposed to closely approximate \eqref{eq:universal-dist}. 
These are, following the MDL literature, also referred to as universal distributions.
We propose approximating $\log \overline p(\DD\,|\, \GG)$ with the \emph{prequential plug-in score} from the prequential plug-in universal distribution, defined as
\begin{align*}
\log p_{\PI}(\DD\,|\,\GG):=\log \prod_{i=1}^n p(x_i \,|\,\hat{\prm}(x_{< i}), \GG),
\end{align*}
where $\hat{\theta}(x_{< i}) \in \Theta_\GG$ indicates a consistent parameters estimate given $x_{<i}:=(x_1,\ldots,x_{i-1})$.

The prequential plug-in score is based on the idea of evaluating a model by its sequential predictive performance and therefore by its generalization capabilities \citep{dawid1999prequential}. The prequential approach in the context of MDL has been proposed by \cite{grunwald2004tutorial,hutter2005asymptotics,grunwald2007minimum}. 
There are advantages in using the prequential plug-in score w.r.t. other scores derived from popular and well-studied universal distributions, such as the log-marginal likelihood (also called \emph{Bayesian score}) $\log p_\text{Bayes}(\DD\,|\,\GG):=\log \int_{\Theta_\GG} p(\DD\,|\, \prm, \GG) p(\prm\,|\,\GG) d\prm$, the \emph{log-normalized maximum likelihood} $\log p_\text{NML}(\DD \,|\,\GG):=\log p(\DD \,|\, \hat{\prm}^\text{MLE}(\DD), \GG)-\log \int_{\ZZ \in {\mathcal X}^n} p(\ZZ\,|\,\hat{\prm}^\text{MLE}(\ZZ), \GG) d\ZZ$ \citep{rissanen1996fisher}, or other approximations. The prequential plug-in score is better suited to neural networks than the Bayesian score, as it does not require integration over the parameters. Whilst it might appear that these two scores imply different preferences for model selection, they are equivalent for several, and often natural, choices of $p(\prm)$ and $\hat{\prm}(x_{< i})$ (see \secref{sec:PPIT}). The log-normalized maximum likelihood is widely used in theoretical treatments of MDL. However, the normalization term over all possible observed data makes this score often intractable or not defined. The well-known AIC/BIC scores can also be cast as approximations to $\log \overline p(\DD|\GG)$ \citep{lam1994learning}, but both are known to have poor empirical performance \citep{silander2008factorized} as they can be quite loose.

\paragraph{Decomposability over \cpds.} 
The assumption that each \cpd~is modelled by a separate set of parameters enables us to write the prequential plug-in score as
\begin{align}
\log p_{\PI}(\DD \,|\, \GG) &=\sum_{d=1}^D\sum_{i=1}^n \log p(x_i^d \,|\, \pa(x_i^d), \hat{\prm}^d(x_{<i})),
\label{eq:PrequentialMDLDecomposed}    
\end{align}
where $\pa(x^d_i)$ indicates the observed values of $\pa(X^d)$ for observation $x_i$ and $\hat{\prm}^d(x_{<i})$ the parameters learned using $\{(x_j^d, \pa(x_j^d))\}_{j=1}^{i-1}$. This decomposition allows a computationally more efficient ranking of structures---for example there are 29,280 DAGs with 5 nodes but only $80$ underlying \cpds.

\subsection{Implementation of the Prequential Plug-In Score with Neural Networks}
\label{sec:preq-nn}
The computation of the prequential plug-in score (\ref{eq:PrequentialMDLDecomposed}) requires evaluating $\log p(x_i^d|\pa(x_i^d), \hat{\prm}^d(x_{<i}))$ $\forall i=1,\ldots,n$. When the \cpds~are modelled by neural networks, we must train the networks to convergence on many subsets of $x_{<i}$ using a stochastic gradient-based optimizer. Modern, usually overparametrized, neural networks 
1) may overfit severely for small $i$, and 
2) training them from scratch for each $i$ can be computationally infeasible, while at the same time it is difficult to use 
them in online settings where the training set constantly grows (a topic of active research). For example, using the model parameters from training on $x_{<i}$ as a starting point for learning model parameters from $x_{<i+j}$ often leads to significantly reduced generalization \citep{ash2020on}.

To overcome the second obstacle, we propose to use the approach described by \citet{blier2018description,bornschein2020small}, specifically to choose a set of increasing split points $\{s_k\}_{k=1}^K$, with $s_k\in [2, \dots, n]$, $s_K{=}n{+}1$ and compute the score of the data 
between two split points $x^d_{s_k}, \dots, x^d_{s_{k+1}-1}$ with a neural network trained from scratch on $x_{<s_k}$ to convergence, which corresponds to the approximation 
\begin{align}
\sum_{i=1}^n \log p(x_i^d | \pa(x_i^d), \hat{\prm}(x_{<i}))\approx \sum_{k=1}^{K-1} \sum_{j=s_k}^{s_{k+1}-1} 
\log p(x_j^d | \pa(x_j^d), \hat{\prm}^d(x_{<s_k})).
\nonumber
\end{align}
In the experiments, we chose the split points to be exponentially spaced 
and performed $K{-}1$ independent training and evaluation runs, usually in parallel.

To overcome the first obstacle, we propose to use a simple confidence calibration approach introduced by \cite{guo2017} to independently calibrate every CPD on every training run. First, consider a network with a softmax output layer for categorical prediction. Conceptually, we could perform post-calibration by first training the network to convergence and then, with all parameters frozen, replacing the output layer $\text{softmax}(h)$ with the calibrated output layer $\text{softmax}(\beta \cdot h)$, where $\beta$ is a scalar parameter chosen to minimize the loss on validation data. In practice, we optimize $\beta$ by gradient descent in parallel with the other model parameters. We alternate computing ten gradient steps for $\theta$, calculated from the training set and using the uncalibrated network (with final layer $\text{softmax}(h)$), with a single gradient step on $\beta$, calculated from the validation set using the calibrated network (with final layer $\text{softmax}(\beta \cdot h)$). This simple calibration procedure has proven to be surprisingly effective at avoiding overfitting when training large neural networks on small datasets \citep{bornschein2020small}. To the best of our knowledge, an analogous method to calibrate continuous-valued neural network outputs does not exist. Thus, we approximate networks for a continuous random variable by networks for a categorical random variable on the quantized values.

\section{Experiments}
\label{sec:Exp}
We demonstrate the effectiveness of our approach, which we refer to as \emph{prequential scoring}, on a variety of synthetic datasets and on a real-world dataset. 

\begin{figure}[t]
\includegraphics[width=0.49\linewidth]{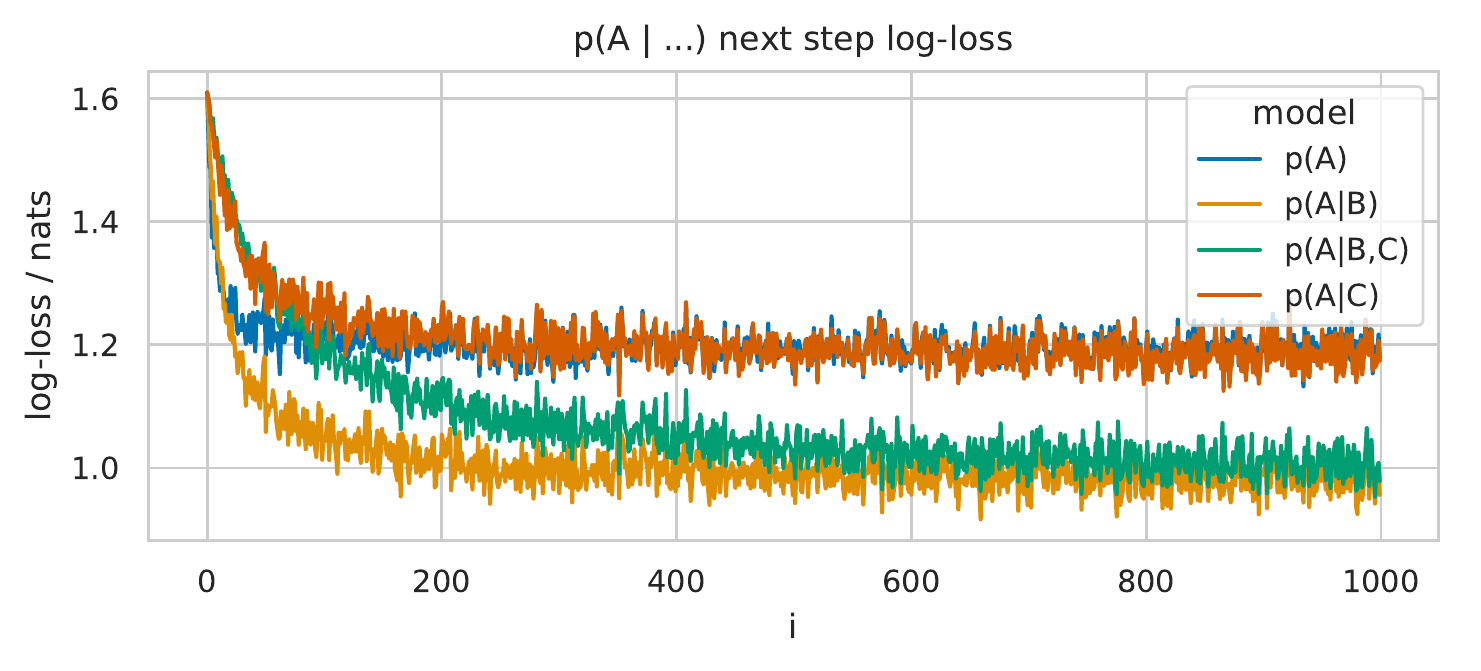}
\includegraphics[width=0.49\linewidth]{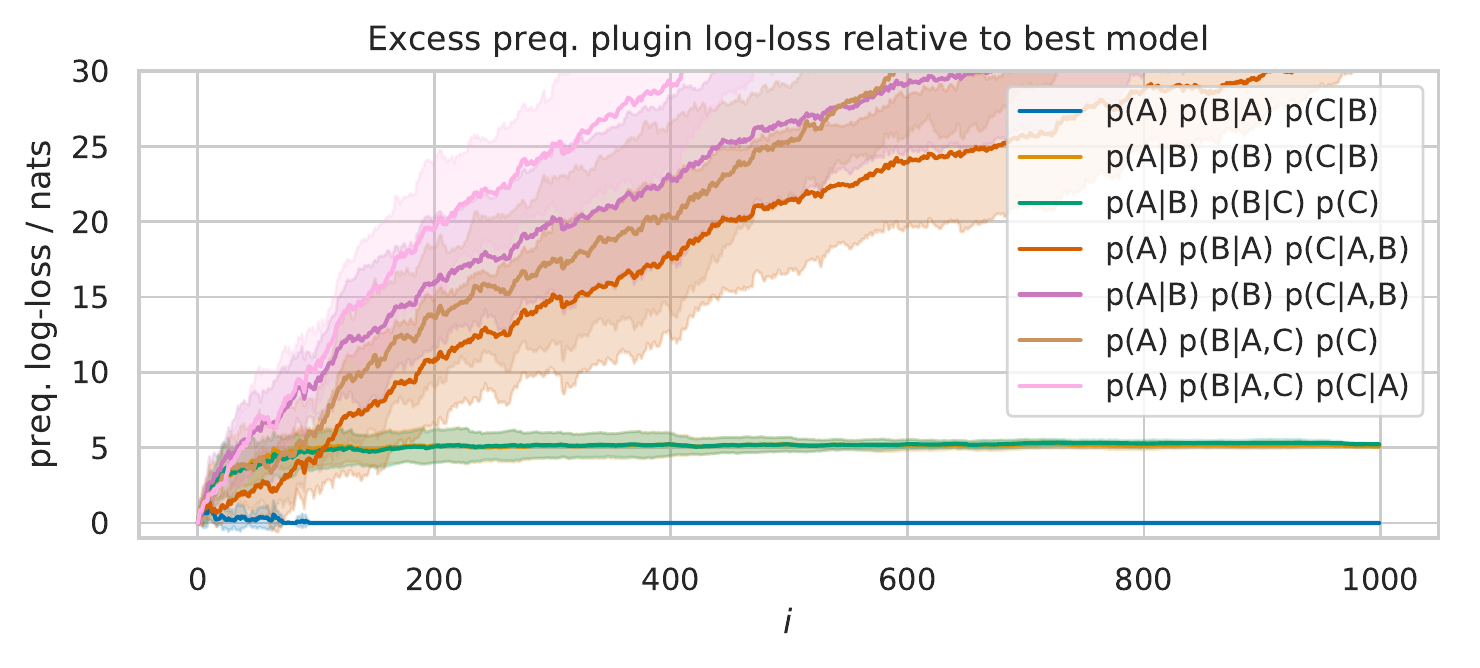}
\vspace{-0.3cm}
\caption{Prequential scoring with tabular \cpds~on synthetic data generated using $\GG^*=A \rightarrow B \rightarrow C$. 
{\bf (a)}: Next-step log-loss for variable $A$ given all possible combinations of the other variables as parents.
{\bf (b)}: Excess prequential log-loss for all possible DAGs relative to $\GG^*$. 
Uncertainty bands show standard deviation over 1,000 permutations of the data.}
\label{fig:tabular}
\end{figure}

\subsection{Prequential Scoring with Tabular \cpds}
\label{sec:PPIT}
Whilst our motivation for introducing the prequential plug-in score is its suitability when neural networks are used to model the \cpds, we first build intuition by considering the case of categorical data with conditional probability tables, i.e. with $p(\XX^d=k\,|\,\pa(X^d)=l,\prm^d)=\prm^d_{k,l}$, which does not require approximating the score.

We generated synthetic data $\DD=\{x_i\}_{i=1}^n$ by using the DAG $\GG^*=A\,{\rightarrow}\,B\, {\rightarrow}\,C$ with each variable taking five possible values, and by drawing the parameters for each \cpd~and value of parents from a Dirichlet distribution with $\alpha^*=1$.
We then computed the \emph{next-step log-loss} $-\log \, p(x_i^d \,|\,\pa(x_i^d), \hat{\prm}^d(x_{<i}))$, $\forall i=1,\ldots,n$ and $\forall d=1,\ldots,D$ for all possible parents sets $\pa(x_i^d)$, using the $\alpha=0.5$-regularized MLE estimator $\hat{\prm}^d_{k,l}(x_{< i}) = \sfrac{N^d_{k,l} + \alpha}{\sum_m (N^d_{m,l} + \alpha)}$, where $N^d_{k,l}$ denotes the number of times that, in $x_{<i}$, $\XX^d$ and $\pa(\XX^d)$ take values $k$ and $l$ respectively.

\figref{fig:tabular}(a) displays the next-step log-loss $\forall i=1,\dots,n$ for variable $A$ given all possible parents sets, averaged over $1,000$ different permutations of the 
datapoints to make the plot less noisy. This average can be seen as an approximation to 
the {\em generalization log-loss} $-\E{\tilde{x} \sim p^*}{\log p(\tilde{x}^d\,|\,\pa(\tilde{x}^d), \hat{\prm}^d(x_{<i}))}$ (i.e. the negative log-likelihood on held-out data). The plot shows that conditioning $A$ on $B$ generally gives the best result. 
Additionally conditioning on $C$ reaches the same performance when sufficient training data is available, 
but results in worse performance in the small-data regime.
In other words, if we were to train on e.g. $1,000$ datapoints and use the generalization log-loss to select a model, we would not be able to reliably select $p(A\,|\,B)$ over $p(A\,|\,B, C)$. The generalization log-loss does not account for model complexity and might lead us to select models that are more complex than necessary. 
However, with only $100$ training examples this loss does give a clear signal that we should prefer $p(A\,|\,B)$, because the 
over conditioned $p(A\,|\,B, C)$ has significantly worse performance in the small-data regime. 
These observations suggest that we should select models in the small-data regime, but finding the right regime could be difficult. \figref{fig:tabular}(a) indicates that the regime is between $\approx 50$ and $\approx 200$ for $p(A|\cdot)$, but that is not known a-priori. Additionally, the optimal regime to perform model selection for e.g. $p(B|\cdot)$ might be different. By summing the next-step log-losses up to $i$, the \emph{prequential log-loss} $-\log\, p_\text{\PI}(x_{\le i}\,|\,\GG)$ accumulates and persists the differences from the small-data regimes. \figref{fig:tabular}(b) shows the prequential log-loss $\forall i=1,\dots,n$ for all DAGs relative to the best one. 
The ground-truth DAG $A\,{\rightarrow} \,B\, {\rightarrow}\,C$ is identified as most plausible, followed by $A\, {\leftarrow} \,B\, {\rightarrow} \,C$ and $A\,{\leftarrow} B\, {\leftarrow} \,C$; all three are in the same Markov equivalence class. The fully connected DAGs, which reach about same next-step log-loss after being trained on 
${\approx}1,000$ datapoints, accumulate more than 50 nats additional loss compared to $A\,{\rightarrow}\,B\, {\rightarrow}\,C$. Notice that with our choice for the parameters estimator the prequential log-loss becomes equivalent to the log-marginal likelihood with $\prm^1,\ldots,\prm^D$ independent random variables with Dirichlet distributions (see Appendix \ref{appsec:relation}).

\subsection{Prequential Scoring with Neural Networks}
\label{sec:ExpNN}
We evaluate \method{} with neural networks on several synthetic datasets and on the Sachs real-world dataset \citep{sachs2005causal}.
We primarily compare with the DAG-GNN method introduced by \cite{yu2019dag}, which represents one of the more competitive modern methods 
to structure learning with neural networks. Additionally we compare with the PC algorithm, a constraint-based method using linear-regression 
based Pearson correlation as independence test \citep{spirtes2000causation}\footnote{Constraint-based approaches use independence tests to infer the existence of links between pairs of variables and require faithfulness---see Appendix \ref{sec:appMarkov} for a discussion on this assumption.}.

\paragraph{Architecture and Hyperparameters.}
In all experiments, we modelled the \cpds{} with neural networks consisting of 3 fully connected layers of 512 hidden units, ReLU activation functions, and dropout with probability of $0.5$ on all the hidden units. We applied a random Fourier transformation to the data obtained by sampling 512 random frequencies from a Gaussian distribution $\NN(0, 10^2)$, as this has been shown to improve the performance in neural networks with low-dimensional inputs \citep{tancik2020fourfeat}. To use the softmax confidence calibration described in \secref{sec:preq-nn}, we mapped the predicted values into the interval $[-1, 1]$ with $\tanh$ and then discretized them according to a uniform 128-values grid.
For optimization, we used Adam \citep{kingma2015adam} with a batch size of 128; for each point $s_k$, we independently choose the learning rate to be either
$1 {\cdot} 10^{-4}$ or $3 {\cdot} 10^{-4}$ depending on which one resulted in a lower calibrated log-loss. We performed 25,000 gradient steps but used early-stopping if the calibrated log-loss increased, which led to considerable compute savings as many training runs on small subsets 
of the data converge after only a few hundred or thousand gradient steps. 
All experiments were carried out on CPUs without accelerators as the networks were relatively small, and a typical run to convergence took between 5 and 15 minutes.
We performed such training runs for each potential \cpd\, and for $K \approx 6$ log-equidistantly spaced 
split-points $s_k$. For example, with 5 observed variables we ran 960 independent training runs corresponding to all combinations of the 80 \cpds, 6 split-points, and 2 learning-rates. We collected and accumulated the results 
and performed an exhaustive search over all DAGs to find the most likely one given the data.
Changing the depth and width of the networks did not impact the rankings of the structures,
provided that the models had sufficient capacity. Similarly, changing the optimizer to RMSprop \citep{graves2014generating} or Momentum SGD \citep{qian1999on} had minimal effect.
This robustness, together with the fact that \method{} has no hyperparameter for regulating the sparsity of the inferred graphs, allowed us to use the same hyperparameter settings throughout all the experiments.

\paragraph{Case Studies for 3 and 5-Node DAGs.}
\label{sec:one_synthetic_model}
\begin{figure}
\includegraphics[width=0.49\linewidth]{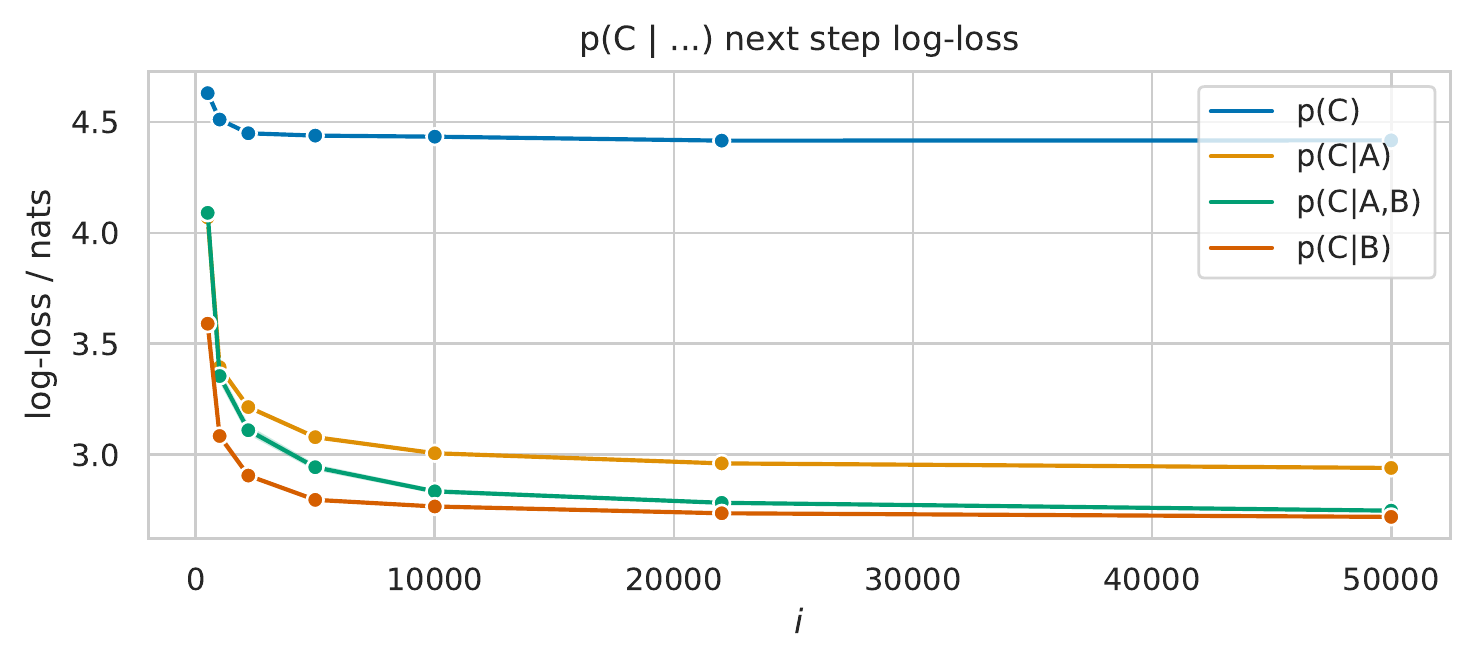}
\includegraphics[width=0.49\linewidth]{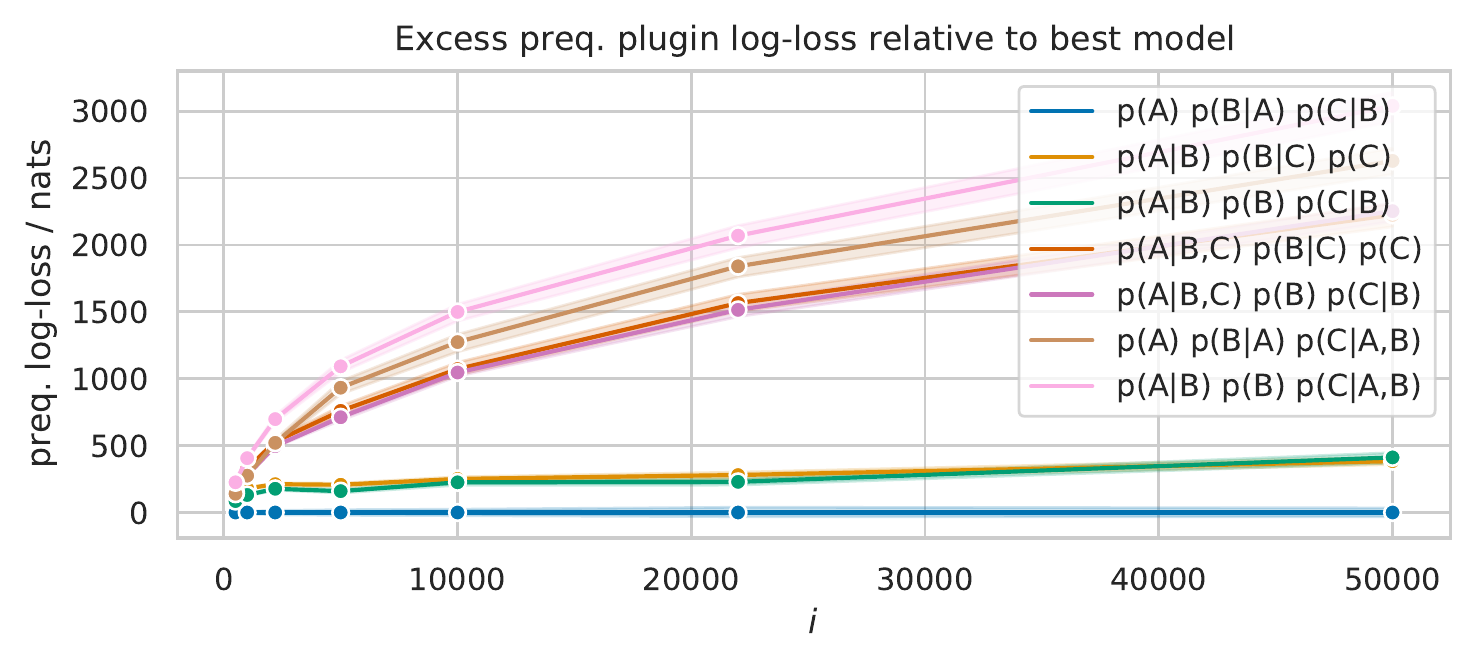}
\caption{Prequential scoring with neural networks on synthetic data generated from $\GG^*=A \rightarrow B \rightarrow C$.
{\bf (a):} Next-step log-loss for variable $C$ given all possible combinations of the other variables as parents.
{\bf (b):} Excess prequential log-loss for all possible DAGs relative to $\GG^*$.
Uncertainty bands show standard deviation over 5 random seeds.
}
\label{fig:3chain-nn}
\end{figure}
\begin{figure*}[t]
\includegraphics[width=\textwidth]{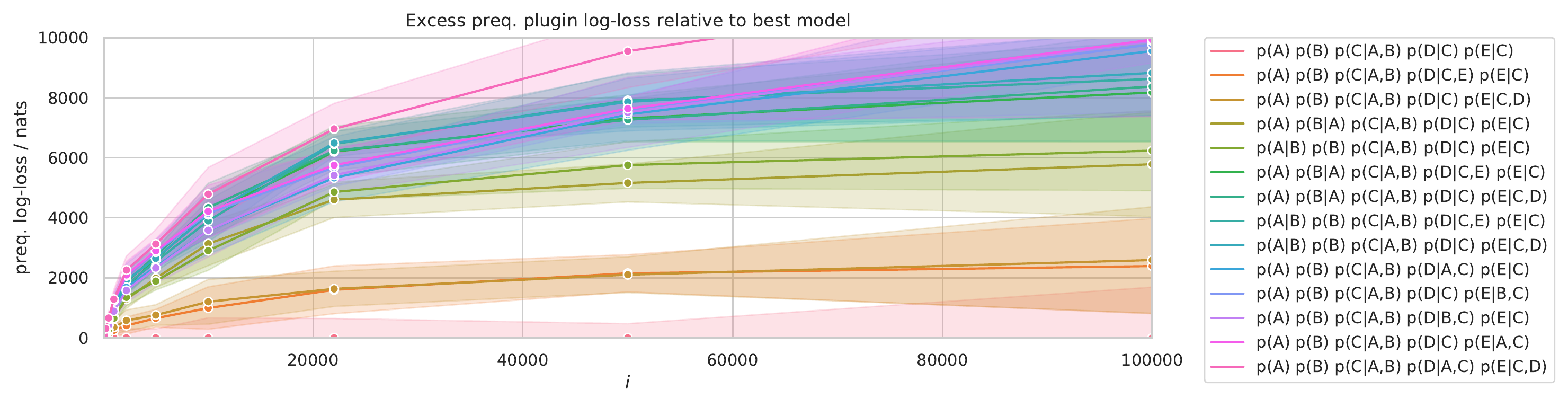}
\caption{Prequential scoring with neural networks on synthetic data generated from $\GG^*=A \rightarrow C \leftarrow B, C \rightarrow D, C \rightarrow E$.
Excess prequential log-loss for the 14 most likely DAGs relative to the   %
the most likely one. Uncertainty bands show standard deviation over 3 random seeds for neural network initialization and mini-batching.}
\label{fig:5star-mdl}
\end{figure*}

We first evaluated \method{} on data generated from hand-crafted generation mechanisms. Below, we describe the results from two mechanisms (a third mechanism is reported in Appendix \ref{appsec:5chain}). \figref{fig:3chain-nn} show the results obtained on data generated from the DAG $\GG^*=A \rightarrow B \rightarrow C$ with $A \sim \NN(0, 1)$, $B = \sin(A + \epsilon_B)$, $C = \sin(B + \epsilon_C)$, and $\epsilon_B,\epsilon_C\sim \NN(0, 0.1^2)$.
We observe the same scaling behaviour of \secref{sec:PPIT} for both the next-step log-loss and the prequential log-loss, and the retrieval of $\GG^*$ with an almost 500 nat margin. \figref{fig:5star-mdl} shows the results obtained on data generated from $\GG^*= A \rightarrow C \leftarrow B, C \rightarrow D, C \rightarrow E$ with $A \sim \NN(0, 1),~B \sim \NN(0, 1),~C = \sin(2AB + \epsilon_C), ~D = \sin(C) + \epsilon_D, ~E = \sin(3C + \epsilon_E)$, and $\epsilon_C,\epsilon_D,\epsilon_E \sim \NN(0, 0.1^2)$.
As above, we observe that the next-step log-loss accumulated on small and medium-sized subsets of the dataset (smaller than roughly 20,000) is crucial to getting a discerning signal for DAG selection. Once again, \method{} identifies $\GG^*$ by a significant margin of 2,000 nats. 

\paragraph{Data from \cite{yu2019dag}.}
\label{sec:ExpYue}
\begin{table*}[t]
\centering
\scalebox{0.65}{
\input{table-daggnn-comparison}
}
\caption{Results on nonlinearities from \cite{yu2019dag}. We list the ground-truth DAGs and the DAGs inferred by DAG-GNN and by \method. 
We also report whether the inferred DAGs are in the Markov equivalence class ${\mathcal E}$ of the ground-truth DAGs and the structural Hamming distance ${\mathcal H}$. }
\label{table:DAG-GNN_comparison}
\end{table*}
To test \method{} on a larger gamut of distributions, we turned to the data generating mechanism introduced by \cite{yu2019dag} for validating DAG-GNN. 
Specifically, we generated datasets by the fixed points $X$ of the equations
\emph{a)} $X = A^\top\cos(X + {\bf 1}) + Z$ and \emph{b)} $X = 2\sin(A^\top(X+0.5\cdot {\bf 1}))+ A^\top(X+0.5\cdot{\bf 1}) + Z$,
where ${\bf 1}$ denotes the all-ones vector and $Z$ a standard normal variable. Due to the limited scalability of the exhaustive DAG search step required in \method, we restricted ourselves to 5$\times$5 adjacency matrices $A$. We generated 5 datasets using \emph{a)} and 5 datasets using \emph{b)}.

The structures inferred by \method{} and DAG-GNN are given in Table~\ref{table:DAG-GNN_comparison}. 
Prequential scoring tends to recover DAGs with lower structural Hamming distance $\mathcal H$ to the ground-truth DAG (0.7 vs 1.3 on average) and that are more frequently in the same Markov equivalence class ${\mathcal E}$ (8 of 10 vs 6 of 10). 
The PC algorithm also performs reasonably well by inferring DAGs within the correct 
Markov equivalence class in 5 cases (details are given in Appendix \ref{appsec:daggnn-pc}). 
Inspection of the data revealed that the vast majority of relationships between the observed variables can be well approximated with simple linear functions; as such, it is not surprising that the PC algorithm performs well even though it uses linear-regression based Pearson correlation as independence test. 

\paragraph{Complex Nonlinearities from Polynomials and Trigonometric Functions.}
\label{sec:ExpOur5Nodes}
\begin{table}[h!]
\centering
\input{table-5nodes}
\vskip0.2cm
\caption{Results obtained with \method{} on 20 randomly generated five-node graphs, with \cpds~corresponding to compositions of polynomials and trigonometric functions (see Table~\ref{table:nonlinear.generation} in Appendix \ref{sec:appOurNonlinearity}). We list the ground-truth DAGs, the inferred DAGs, the structural Hamming distance ${\mathcal H}$, and whether the inferred DAGs are in the Markov equivalence class ${\mathcal E}$ of the ground-truth DAGs.}
\label{table:OurNonlinearity}
\end{table}
To investigate the performance of \method{} for setting in which the \cpds~need to model highly nonlinear relationships, 
we created a potpourri of synthetic data by first sampling random 5-node DAGs using the GNP algorithm \citep{batagelj2005efficient} with link probability $0.25$ and then annotating the links with random compound functions of polynomials, trigonometric functions, reciprocals and random noise. We created a set of 20 such data generation mechanisms which are listed in Table~\ref{table:nonlinear.generation} of 
Appendix~\ref{sec:appOurNonlinearity}. For example, a typical generation mechanism is $A=\sin(30\, \epsilon_A)$, $B=\sin(2A)+\epsilon_B$, $C=\sin(B^3-A+\epsilon_C)$, 
$D=(C+\epsilon_D)^3$, $E=\sign(A)(|A|+0.1)^{-1}+\epsilon_E$, with $\epsilon_A, \epsilon_B, \epsilon_C, \epsilon_D, \epsilon_E\sim \NN(0, 0.1^2)$.

In Table~\ref{table:OurNonlinearity} we list the ground-truth DAGs, the DAGs inferred by \method{}, the structural Hamming distance ${\mathcal H}$, and whether the inferred DAGs are in the Markov equivalence class ${\mathcal E}$ of the ground-truth DAGs.
We observe that \method{} infers DAGs with an average ${\mathcal H}$ of 1.9 and recovers a member of ${\mathcal E}$ in 12 of the 20 cases.

We applied DAG-GNN and PC to the same 20 data datasets and were not able to obtain reliable and reproducible results. For DAG-GNN with default hyperparameters, the inferred structure varied significantly for different initial seeds and different sizes of the dataset. We also noticed a high sensitivity to the sparsity regularization hyperparameter. For the constraint-based method the results appear 
random. We did however not expect reliable results because the independence test is not designed to work on strongly nonlinear data. 

\paragraph{Protein Signaling Network.}
\begin{wrapfigure}[15]{r}{0.4\textwidth}
\vspace{-0.8cm}
\scalebox{0.8}{
\begin{tikzpicture}
\Vertex[x=3.0, y=8.127906061693181e-09, label=pip3, position=below]{pip3}
\Vertex[x=2.523760559755236, y=1.621922313506374, label=plcg, position=above]{plcg}
\Vertex[x=1.2462452447075993, y=2.728895895974767, label=pip2, position=above]{pip2}
\Vertex[x=-0.4269444905793424, y=2.9694642941005434, label=pkc, position=above]{pkc}
\Vertex[x=-1.964581999103687, y=2.2672489243894876, label=pka, position=above]{pka}
\Vertex[x=-2.8784789124484558, y=0.845197770567386, label=raf, position=above]{raf}
\Vertex[x=-2.8784789124484558, y=-0.8451976649046071, label=mek, position=below]{mek}
\Vertex[x=-1.9645821779176205, y=-2.2672487293197423, label=erk, position=below]{erk}
\Vertex[x=-0.4269450270211425, y=-2.9694642778447315, label=akt, position=below]{akt}
\Vertex[x=1.246245334114566, y=-2.728895879718955, label=p38, position=below]{p38}
\Vertex[x=2.5237603809413027, y=-1.621922654878429, label=jnk, position=below]{jnk}
\Edge[Direct, color=black](pip3)(pip2)
\Edge[Direct, color=blue, style=dotted, ](pip3)(akt)
\Edge[Direct, color=blue, style=dotted, ](pip3)(plcg)
\Edge[Direct, color=black](plcg)(pip2)
\Edge[Direct, color=blue, style=dotted, ](pip2)(pkc)
\Edge[Direct, color=black](pkc)(pka)
\Edge[Direct, color=black](pkc)(mek)
\Edge[Direct, color=blue, style=dotted, ](pkc)(raf)
\Edge[Direct, color=black](pkc)(jnk)
\Edge[Direct, color=black](pkc)(p38)
\Edge[Direct, color=red, style=dashed, ](pkc)(plcg)
\Edge[Direct, color=black](pka)(akt)
\Edge[Direct, color=black](pka)(erk)
\Edge[Direct, color=blue, style=dotted, ](pka)(mek)
\Edge[Direct, color=blue, style=dotted, ](pka)(raf)
\Edge[Direct, color=blue, style=dotted, ](pka)(jnk)
\Edge[Direct, color=blue, style=dotted, ](pka)(p38)
\Edge[Direct, color=blue, style=dotted, ](raf)(mek)
\Edge[Direct, color=blue, style=dotted, ](mek)(erk)
\Edge[Direct, color=red, style=dashed, ](mek)(raf)
\Edge[Direct, color=red, style=dashed, ](mek)(p38)
\Edge[Direct, color=red, style=dashed, ](mek)(akt)
\Edge[Direct, color=blue, style=dotted, ](erk)(akt)
\Edge[Direct, color=red, style=dashed, ](akt)(erk)
\Edge[Direct, color=red, style=dashed, ](p38)(plcg)
\Edge[Direct, color=red, style=dashed, ](p38)(jnk)
\end{tikzpicture}}
\end{wrapfigure}
As real-world dataset we considered the Sachs dataset for the discovery of a protein signaling network \citep{sachs2005causal}, a benchmark dataset for structure learning with experimental annotations accepted by the biology research community. The data contains continuous measurements of expression levels of multiple phosphorylated proteins and phospholipid components in human immune system cells, and the network provides the ordering of the connections between pathway components. Based on
$n = 7,466$ samples of $D = 11$ cell types, \cite{sachs2005causal} estimated 20 links in the graph. 
In addition to observational samples, the dataset contains interventional samples obtained by activating or inhibiting expression levels at particular nodes. We handled interventional samples following the principles behind intervention in causal Bayesian networks \citep{pearl2000causality,pearl2016causal}: if sample $x_j^d$ was marked as the result of an intervention, we did not consider it for the learning and evaluation of the $d$-th \cpd~(see Appendix~\ref{sec:tabularInterventions} for a description and a visualization of the effect that interventional data can have on \method{}).

We computed the prequential plug-in score from three runs with different random seeds for neural network initialization and mini-batching. For scalability reasons, we were only able to consider \cpds~with at most 4 parents, and we used a heuristic hill-climbing method \citep{heckerman1995learning} to search the space of DAGs instead of an exhaustive search. 

\begin{wrapfigure}[7]{l}{0.47\textwidth}
\vspace{-0.15cm}
\begin{tabular}{|c | c | c|}
\hline
 & ${\mathcal H}$ & \# links \\
\hline
NOTEARS & 22 & 16\\
DAG-GNN & 19 & 18\\
\method{} & 16 & 15\\
prequential scoring (PWA) & 18.4 & 16.8 \\
\hline
\end{tabular}
\end{wrapfigure}

Above, we show the DAG $\GG^*$ that has been accepted by the biology research community as the best known solution overlayed with the DAG inferred by \method{}. Shared links are solid in black; otherwise, links discovered by \method{} are dotted blue and the DAG links only in $\GG^*$ are dashed in red. The table on the left reports the structural Hamming distance ${\mathcal H}$ and the number of links found by prequential scoring, DAG-GNN and NOTEARS \citep{zheng2018dags}. \Method{} performs favourably and finds a graph with lower structural Hamming distance ${\mathcal H}$ to the ground-truth compared to DAG-GNN and NOTEARS.

However, \method{} identified a number of DAGs with good prequential plug-in scores and partially overlapping uncertainty bands. 
Using Bayes rule we can approximate the posterior $p(\GG | \DD)$ and compute the posterior weighted average (PWA) Hamming distance
$\sum_\GG p(\GG | \DD) {\cal H}(\GG, \GG^*)$ and the posterior weighted number of links.
To approximate the posterior we considered all graphs $\GG$ visited by the hill climbing algorithm. The results are reported in the table.
The posterior was dominated by a few hundred graphs $\GG$ with good scores.

\section{Discussion}
\label{sec:discussion}
This paper considered the problem of learning the structure of a Bayesian network in settings in which modern, possibly overparametrized, neural networks might be used to model its conditional distributions. 
We proposed the use of the prequential plug-in score as a MDL-based model selection criterion that does not require any explicit sparsity regularization, and provided a specific implementation using neural networks that shows good performance, leads to sparse, parsimonious structural inferences, and often recovers the structure underlying the data generation mechanism. 

Previous literature on MDL-based approaches to structure learning has focused on analytically tractable model families, such as tabular distributions 
or distributions with conjugate priors \citep{grunwald2007minimum}. 
For categorical random variables, \cite{silander2008factorized} derived a factorized approximation 
to the normalized maximum likelihood by exploiting the conditional distribution structure. 
This work was further extended \citep{silander2018quotient} to focus on 
 model selection procedures that assign the same score to every model in a Markov equivalence class. 
However, it is not clear how to generalize these techniques to continuous random variables or complex models.

Outside the context of structure learning, MDL-inspired model selection for neural networks has primarily focused on using variational approximations or approximations based on AIC or BIC \citep{hinton1993keeping,mackay2003information}. 
\cite{lehtokangas1993neural} were perhaps the first to use a prequential plug-in approach, though the structure learned was the capacity of the neural network. \cite{blier2018description} pioneered using the prequential plug-in distribution for modern scale neural networks architectures and essentially argued that prequential coding leads to much shorter description lengths than state-of-the-art variational approximations.
They used the block-wise estimates described in \secref{sec:preq-nn} but without the confidence calibration; as a result, they had to switch between different model classes to avoid overfitting. They thus calculated prequential MDL 
estimates for a particular switching pattern, not for a model class. \cite{bornschein2020small} extended the block-wise estimate with calibration and obtained MDL estimates for modern overparametrized neural networks without limiting their capacity.

Recent efforts on structure learning with modern neural networks has focused on improving scalability by framing structure search as a continuous optimization problem with regularized maximum-likelihood as a scoring metric (see  \citet{zheng2018dags,yu2019dag,zheng2020learning,pamfil2020dynotears}, and \citet{vowels2021dya} for a review). 
Scalability is an important aspect that we did not consider. As a consequence, our experiments were only feasible with a small number of variables. 
While this might seem like a step backwards compared to recent work, we believe that it is important to investigate
new scoring metrics without the confounding effect of approximating the search procedure. 
Proposals for scaling prequential scoring to a higher number of variables include
classical approximation methods developed for Bayesian scores \citep{heckerman1999tutorial}, 
techniques like dynamic programming \citep{malone2011memory}, branch and bound \citep{campos2011efficient}, 
mathematical programming \citep{jaakkola2010learning,cussen2011bayesian}, and continuous optimization approaches  \citep{zheng2018dags,yu2019dag,zheng2020learning,pamfil2020dynotears}.

Our approach effectively uses the generalization performance when trained on limited data as a model selection criterion. As such it is related to recent work that uses adaptation speed, i.e. how quickly models adapt to changes in the data generating process, to infer causal structures \citep{ke2019learning,bengio2020a} (see \citet{lepriol2020analysis} for a theoretical justification of this principle).
The prequential MDL perspective offers an alternative and potentially simpler argument based on sample-efficiency instead of gradient-step efficiency to justify such an approach. And, as we have shown, this perspective is not only theoretically well-motivated but also applicable to i.i.d. data. %

\bibliography{bib}

\newpage

\appendix
\appendix
\section{Notes on MDL}
Given a sufficiently regular $k$-dimensional parametric model class $\MM_\GG$ (e.g.\ a curved exponential family), an essential result from the MDL literature is that $\log p_\text{Bayes}(\ZZ\,|\,\GG)$ and $\log p_\text{plug-in}(\ZZ\,|\,\GG)$ closely track the maximum log-likelihood of $\MM_\GG$ for all sufficiently regular data $\ZZ\in {\mathcal X}^n$ (e.g.\ the MLE lies in the interior of the parameter space); see \cite{grunwald2019minimum}[Section 2.2] for precise statements and pointers to the literature. Essentially, this result shows that $p_\text{Bayes}$ and $p_\text{plug-in}$ are universal distributions and satisfy \eqref{eq:universal-dist}, up to a small error.%
\begin{theorem}
\label{thm:universal.forecaster}
Under sufficient regularity conditions on the model class $\MM_\GG$ and for any sufficiently regular data $\ZZ\in {\mathcal X}^n$,
\begin{align*}
     \log p(\ZZ\,|\,\hat\theta^{\text{MLE}}(\ZZ),\GG)  
     &\leq
     \log p_\text{\PI}(\ZZ\,|\,\GG) + \OO(\log(n)) + \OO(1),\\
    \log p(\ZZ\,|\,\hat\theta^{\text{MLE}}(\ZZ),\GG)  
     &\leq
     \log p_\text{Bayes}(\ZZ\,|\,\GG) + \frac{k}{2}\log(n) + \OO(1),
 \end{align*}
where the constants depend on the mismatch between $\MM_\GG$ and the data distribution.
\end{theorem}

\paragraph{Connection to Data Compression.}
The use of universal distributions connects to data compression through the Kraft–McMillan inequality \citep{cover1999elements}, which states that a probabilistic model's log-loss on $\DD$ is equal to the shortest achievable code-length for encoding $\DD$ using a code derived from that model. However, we cannot use the optimal choice $\hat\theta^{\text{MLE}}(\DD)$ for compression since it is not known a priori. Instead, we want a code that can losslessly encode almost as well as $p(\cdot|\hat\theta^{\text{MLE}},\GG)$ by having a code length near $\min_\prm -\log \, p(\mathcal Z | \prm, \GG)$ for all possible data $\ZZ$. A reader may recognize this criterion as \eqref{eq:universal-dist}, so we can alternatively define a universal distribution as one that achieves a code length as closely to that of $p(\cdot|\hat\theta^{\text{MLE}},\GG)$ as possible for all data $\ZZ$.

\section{Faithfulness Example}
\label{sec:appMarkov}
Let us denote with $\II(\GG)=\{X \indep_{\GG} Y \,|\, Z\}$ the set of statistical independence relationships implied by the structure of a BN $(\GG,p)$, also called \emph{global Markov independencies}, and let us denote with $\II(p)$ the set of 
statistical independence relationships satisfied by the distribution $p$. We must have $\II(\GG)\subseteq \II(p)$, i.e. any independence encoded in $\GG$ is satisfied by $p$---we denote this as $X \indep_{\GG} Y \,|\, Z \Rightarrow X \indep_{p} Y |Z$.
If the converse relation holds, namely if any independence satisfied by $p$ is encoded in $\GG$ (i.e. $X \indep_{p} Y \,|\, Z \Rightarrow X \indep_{\GG} Y \,|\, Z$) giving $\II(\GG)=\II(p)$, then we say that $p$ is \emph{faithful} to $\GG$.
Unfaithfulness can arise for different reasons, for example when dependence along different pathways cancels 
out \citep{peters2017elements} or when links express deterministic relationships, which can occur in many real-world
scenarios (see \cite{koski2012review,mabrouk2014efficient}). 

Constraint-based approaches to structure learning assume faithfulness of the distribution $p$ to the structure $\GG$ underlying the data generation mechanism, and infer a member of the Markov equivalence class w.r.t. which $p$ is faithful. In the example below we shows that, when $p$ is not faithful, recovering the Markov equivalence class w.r.t. which $p$ is faithful might lead to a preference for more complex \cpds. This issue is not shared by \method{} which does not assume faithfulness and infers a structure by taking the complexity of the \cpds~into account.  

\begin{wrapfigure}[4]{r}{0.3\textwidth}
\begin{tikzpicture}[dgraph]
\node at (0,0.4) {${\cal G}$}; 
\node[dot] (A) [label= south:$A$] at (-1.5,0) {};
\node[dot] (C) [label= south:$C$] at (0,0) {};
\node[dot] (B) [label= south:$B$] at (1.5,0) {};
\draw[line width=0.6pt](A)--(C);
\draw[line width=0.6pt](B)--(C);
\draw[->, line width=0.6pt](A)to [bend left=+48] (B);
\end{tikzpicture}
\end{wrapfigure}

Let us consider the data generation mechanism
\begin{itemize}
    \item $A=$ random prime number with $2 \le A \le M$,
    \item $B=$ random prime number with $A \le B \le M$,
    \item $C= A \cdot B$.
\end{itemize}

The DAG $\GG$ corresponding to this mechanism is shown in the figure above and has $\II(\GG)=\emptyset$.

If we analyze the conditional independence relationships satisfied by the joint distribution $p$, we realize that any number $c$ with $p(C=c)>0$ can be uniquely factorized into the numbers $A$ and $B$ that generated
it. This is a direct consequence of the fundamental theorem of arithmetic and the fact that the 
generative process ensures $A\le B$. Because $A$ does not provide additional information about $B$ 
given $C$ and vice versa, the set of statistical independence relationships between the 
variables is given by $\II(p)=\{A \indep B \,|\, C\}$. Therefore $p$ is {\em not} faithful to $\GG$. 

The Markov equivalence class ${\mathcal E}$ of DAGs w.r.t $p$ is faithful is ${\mathcal E}=\{A\,\leftarrow\, C\,\rightarrow\,B,\, A\,\rightarrow \,C\,\rightarrow \,B,\, A\,\leftarrow \,C \,\leftarrow \,B\}$.
The structures in ${\mathcal E}$ can be considered simpler than $\GG$ as they contain fewer links. 
However, their joint distributions must contain at least one conditional distribution which requires an
integer factorization. For example, $p(A \,|\, C)$ reads: "perform a factorization of $C$ into prime numbers and set $A$ to the smaller one". Factorization is considered a hard problem and it is commonly used as a one-way function in cryptography because no known algorithms can perform it efficiently.

Applying structure learning methods that infer the Markov equivalence class w.r.t which the distribution is faithful would return ${\mathcal E}$ and, along the way, the methods would have to discover (the existence of) 
a factorization algorithm (or table) for the natural numbers up to $M^2$. Therefore,
whilst returning simpler structures, these methods would imply significantly more complex conditional distributions (for some intuitive definition of complexity). \Method{} instead takes the complexity of the conditional distributions into account, and therefore it is free to prefer a graph which is outside of $\mathcal E$ if it finds modeling a multiplication "easier" than factorization.

\section{Case Study on 5 Nodes Chains}
\label{appsec:5chain}
In the sections above we argued that \method{} not only takes the overall DAG 
sparsity into account, but also considers the complexity of the \cpds~as measured by their sample-efficiency. As a result, \method{} often has a preference for the direction of links even when the resulting DAGs contain the same number of edges and belong to
the same Markov equivalence class.

In this section, we demonstrate this property by generating data from two chains of 5 variables $X^1,\cdots,X^5$. 
In the first chain, each variable is an almost linear function of the previous variable in the chain plus some noise. 
In the second chain, the functional relationship between successive nodes is more non-linear (periodic) and requires a multi-modal CPD when modeling 
the relationship in the inverse direction. Note that our models are capable of modeling multi-modal regression variables because we use categorical prediction for the discretized target variable as described in \secref{sec:preq-nn}.

For each of the two chains, we show the top-5 inferred DAGs and their excess scores relative to the best DAG. 
We observe that for both chains \method{} correctly identifies the ground-truth DAG as most likely. However, \method{} consistently
avoids inverting the direction of the links in the second chain, which is not a property shared by the contenders.
\vspace{0.5cm}

\scalebox{0.89}{
\begin{tabular}{c|c}
    \multicolumn{2}{c}{$X^1 \sim \NN(0, 1)\; ; \; X^d \sim \NN(\sin(X^{d-1}), 0.1^2)$} 
\\[5pt]
    Inferred DAG & $\Delta \log p(\DD | \GG)$ \\ 
\hline
    \begin{tikzpicture}[dgraph]
    \clip (-.4,-.4) rectangle (4.4,.4);
    \node[dot] (A) at (0,0) {};
    \node[dot] (B) at (1,0) {};
    \node[dot] (C) at (2,0) {};
    \node[dot] (D) at (3,0) {};
    \node[dot] (E) at (4,0) {};
    \draw[->, line width=0.6pt] (A) to (B);
    \draw[->, line width=0.6pt] (B) to (C);
    \draw[->, line width=0.6pt] (C) to (D);
    \draw[->, line width=0.6pt] (D) to (E);
    \end{tikzpicture}
    &
    \wvcentered{ 0 }
\\
    \begin{tikzpicture}[dgraph]
    \clip (-.4,-.4) rectangle (4.4,.4);
    \node[dot] (A) at (0,0) {};
    \node[dot] (B) at (1,0) {};
    \node[dot] (C) at (2,0) {};
    \node[dot] (D) at (3,0) {};
    \node[dot] (E) at (4,0) {};
    \draw[->, line width=0.6pt] (A) to (B);
    \draw[->, line width=0.6pt] (B) to (C);
    \draw[->, line width=0.6pt] (C) to (D);
    \draw[->, line width=0.6pt] (E) to (D);
    \end{tikzpicture}
    &
    \wvcentered{ 1117 }
\\
    \begin{tikzpicture}[dgraph]
    \clip (-.4,-.4) rectangle (4.4,.4);
    \node[dot] (A) at (0,0) {};
    \node[dot] (B) at (1,0) {};
    \node[dot] (C) at (2,0) {};
    \node[dot] (D) at (3,0) {};
    \node[dot] (E) at (4,0) {};
    \draw[->, line width=0.6pt] (B) to (A);
    \draw[->, line width=0.6pt] (B) to (C);
    \draw[->, line width=0.6pt] (C) to (D);
    \draw[->, line width=0.6pt] (D) to (E);
    \end{tikzpicture}
    &
    \wvcentered{ 1178 }
\\
    \begin{tikzpicture}[dgraph]
    \clip (-.4,-.4) rectangle (4.4,.4);
    \node[dot] (A) at (0,0) {};
    \node[dot] (B) at (1,0) {};
    \node[dot] (C) at (2,0) {};
    \node[dot] (D) at (3,0) {};
    \node[dot] (E) at (4,0) {};
    \draw[->, line width=0.6pt] (B) to (A);
    \draw[->, line width=0.6pt] (C) to (B);
    \draw[->, line width=0.6pt] (C) to (D);
    \draw[->, line width=0.6pt] (D) to (E);
    \end{tikzpicture}
    &
    \wvcentered{ 1319 }
\\
    \begin{tikzpicture}[dgraph]
    \clip (-.4,-.4) rectangle (4.4,.4);
    \node[dot] (A) at (0,0) {};
    \node[dot] (B) at (1,0) {};
    \node[dot] (C) at (2,0) {};
    \node[dot] (D) at (3,0) {};
    \node[dot] (E) at (4,0) {};
    \draw[->, line width=0.6pt] (B) to (A);
    \draw[->, line width=0.6pt] (C) to (B);
    \draw[->, line width=0.6pt] (D) to (C);
    \draw[->, line width=0.6pt] (E) to (D);
    \end{tikzpicture}
    &
    \wvcentered{ 1348 }
\end{tabular}
\quad
\begin{tabular}{c|c}
    \multicolumn{2}{c}{$X^1 \sim \NN(0, 1)\; ; \; X^d \sim \NN(\sin(4 \cdot X^{d-1}), 0.1^2)$} 
\\[5pt]
    Inferred DAG & $\Delta \log p(\DD | \GG)$ \\ 
\hline
    \begin{tikzpicture}[dgraph]
\clip (-.4,-.4) rectangle (4.4,.4);
\node[dot] (A) at (0,0) {};
\node[dot] (B) at (1,0) {};
\node[dot] (C) at (2,0) {};
\node[dot] (D) at (3,0) {};
\node[dot] (E) at (4,0) {};
\draw[->, line width=0.6pt] (A) to (B);
\draw[->, line width=0.6pt] (B) to (C);
\draw[->, line width=0.6pt] (C) to (D);
\draw[->, line width=0.6pt] (D) to (E);
\end{tikzpicture}
    &
    \wvcentered{ 0 }
\\
    \begin{tikzpicture}[dgraph]
\clip (-.4,-.4) rectangle (4.4,.4);
\node[dot] (A) at (0,0) {};
\node[dot] (B) at (1,0) {};
\node[dot] (C) at (2,0) {};
\node[dot] (D) at (3,0) {};
\node[dot] (E) at (4,0) {};
\draw[->, line width=0.6pt] (A) to (B);
\draw[->, line width=0.6pt] (B) to (C);
\draw[->, line width=0.6pt] (C) to (D);
\draw[->, line width=0.6pt] (D) to (E);
\draw[->, line width=0.6pt] (C) to [bend left=+18] (E);
\end{tikzpicture}
    &
    \wvcentered{ 2361 }
\\
    \begin{tikzpicture}[dgraph]
    \clip (-.4,-.4) rectangle (4.4,.4);
    \node[dot] (A) at (0,0) {};
    \node[dot] (B) at (1,0) {};
    \node[dot] (C) at (2,0) {};
    \node[dot] (D) at (3,0) {};
    \node[dot] (E) at (4,0) {};
    \draw[->, line width=0.6pt] (A) to (B);
    \draw[->, line width=0.6pt] (B) to (C);
    \draw[->, line width=0.6pt] (C) to (D);
    \draw[->, line width=0.6pt] (D) to (E);
    \draw[->, line width=0.6pt] (B) to [bend left=+18] (D);
    \end{tikzpicture}
    &
    \wvcentered{ 2455 }
\\
    \begin{tikzpicture}[dgraph]
    \clip (-.4,-.4) rectangle (4.4,.4);
    \node[dot] (A) at (0,0) {};
    \node[dot] (B) at (1,0) {};
    \node[dot] (C) at (2,0) {};
    \node[dot] (D) at (3,0) {};
    \node[dot] (E) at (4,0) {};
    \draw[->, line width=0.6pt] (A) to (B);
    \draw[->, line width=0.6pt] (B) to (C);
    \draw[->, line width=0.6pt] (C) to (D);
    \draw[->, line width=0.6pt] (D) to (E);
    \draw[->, line width=0.6pt] (B) to [bend left=+18] (E);
    \end{tikzpicture}
    &
    \wvcentered{ 4569 }
\\
    \begin{tikzpicture}[dgraph]
    \clip (-.4,-.4) rectangle (4.4,.4);
    \node[dot] (A) at (0,0) {};
    \node[dot] (B) at (1,0) {};
    \node[dot] (C) at (2,0) {};
    \node[dot] (D) at (3,0) {};
    \node[dot] (E) at (4,0) {};
    \draw[->, line width=0.6pt] (A) to (B);
    \draw[->, line width=0.6pt] (B) to (C);
    \draw[->, line width=0.6pt] (C) to (D);
    \draw[->, line width=0.6pt] (D) to (E);
    \draw[->, line width=0.6pt] (A) to [bend left=+18] (C);
    \end{tikzpicture}
    &
    \wvcentered{ 4678 }
\end{tabular}}

\section{PC Algorithm on Data from \cite{yu2019dag}}
\label{appsec:daggnn-pc} 
This section describes the behaviour of the PC algorithm on data generated according to the mechanism introduced by \cite{yu2019dag} for validating DAG-GNN described in the main text. We used the implementation in PGMPy \citep{ankan2015pgmpy} with $\chi^2$ ({\tt k2}) and linear-regression based Pearson correlation ({\tt pearsonr}) as conditional independence test. The PC algorithm returns a PDAG, which uses undirected edges to represent the Markov equivalence class, so the Hamming distance to ground-truth is not comparable to the one of \method{}.

\begin{table*}[h]
\scalebox{0.75}{\input{table-daggnn-pc}}
\end{table*}

\section{5-Node DAGs}
\label{sec:appOurNonlinearity}
We applied \method{} to 20 randomly generated DAGs with nonlinear relationships 
between variables. Each instance was created by 1) generating a random 5-node DAG with the GNP algorithm 
\citep{batagelj2005efficient} with link probability of $0.25$, 
2) removing links to make the adjacency matrix lower-triangular, 
3) selecting the non-linear functions for each (topologically sorted) variable uniformly from 
a set of candidate-functions with a the corresponding number of input variables 
(see Table~\ref{table:nonlinear.generation}),
4) generating i.i.d. data (sampling a independent $\epsilon \sim \NN(0, 0.1^2)$ per variable), 

Table~\ref{table:nonlinear.generation} shows the 20 generative processes that were created this 
way. The results when applying \method{} to this data can be found in Table~\ref{table:OurNonlinearity}.
DAG-GNN \citep{yu2019dag} with the recommended parameter settings was unstable with respect to random seed and dataset size.
We were not able to find a parameter setting that worked reliably  , so we are not comfortable %
presenting the performance of DAG-GNN on this data.
\begin{table}[h!]
\scalebox{0.78}{
\begin{tabular}{|l|l|l|l|l|}
\hline
A & B & C & D & E \\
\hline
\hline
$\sin(30 \epsilon)$ & 
$\sin(2A) + \epsilon$ & 
$\sin(B^3 - B + \epsilon)$ & 
$(C + \epsilon)^3$ & 
$\sign(A) \sfrac{1}{|A| + 0.1} + \epsilon$ \\

\hline
$10 \epsilon$ & 
$(A + \epsilon)^3$ & 
$\sin(2A) + \epsilon$ & 
$\sin(\sfrac{1}{|C| + 0.1} + \epsilon)$ & 
$\sin(2A) + \epsilon$ \\

\hline
$10 \epsilon$ & 
$10 \epsilon$ & 
$\sign(B) \sfrac{1}{|B| + 0.1} + \epsilon$ & 
$\sin(2C^3 -B^2) + \epsilon$ & 
$\sign(D) \sin(4DA + \epsilon)$ \\

\hline
$10 \epsilon$ & 
$\sign(A) \sfrac{1}{|A| + 0.1} + \epsilon$ & 
$10 \epsilon$ & 
$\sin(2C^3 -A^2) + \epsilon$ & 
$\sign(D) \sin(4DA + \epsilon)$ \\

\hline
$\sin(30 \epsilon)$ & 
$(A + \epsilon)^3$ & 
$\sin(30 \epsilon)$ & 
$(C + \epsilon)^3$ & 
$\sin(C^3 - C + \epsilon)$ \\

\hline
$10 \epsilon$ & 
$\sin(30 \epsilon)$ & 
$\sin(2B^3 -A^2) + \epsilon$ & 
$\sign(C) \sin(4CA + \epsilon)$ & 
$\sin(\sfrac{1}{|D| + 0.1} + \epsilon)$ \\

\hline
$\sin(30 \epsilon)$ & 
$\sin(A^3 - A + \epsilon)$ & 
$\sin(2B) \sin(\sfrac{1}{|A| + 0.1}) + \epsilon$ & 
$\sin(4CBA + \epsilon$) & 
$\sin(2D^3 -C^2) + \epsilon$ \\

\hline
$\sin(30 \epsilon)$ & 
$\sin(A^3 - A + \epsilon)$ & 
$\sin(2B) \sin(\sfrac{1}{|A| + 0.1}) + \epsilon$ & 
$\sign(A) \sfrac{1}{|A| + 0.1} + \epsilon$ & 
$\sin(2D) \sin(\sfrac{1}{|A| + 0.1}) + \epsilon$ \\

\hline
$10 \epsilon$ & 
$10 \epsilon$ & 
$\sin(4BA + \epsilon)$ & 
$\sin(30 \epsilon)$ & 
$\sin(4DCA + \epsilon)$ \\

\hline
$\sin(30 \epsilon)$ & 
$\sin(A^3 - A + \epsilon)$ & 
$\sin(2B^3 -A^2) + \epsilon$ & 
$10 \epsilon$ & 
$\sin(2B) \sin(4BA + \epsilon)$ \\

\hline
$\sin(30 \epsilon)$ & 
$\sin(30 \epsilon)$ & 
$\sign(B) \sin(2BA + \epsilon)$ & 
$\sin(30 \epsilon)$ & 
$\sin(2D) \sin(4DA + \epsilon)$ \\

\hline
$\sin(30 \epsilon)$ & 
$\sin(2A) + \epsilon$ & 
$\sin(4BA + \epsilon)$ & 
$\sin(2C) \sin(\sfrac{1}{|B| + 0.1}) + \epsilon$ & 
$\sign(C) \sfrac{1}{|C| + 0.1} + \epsilon$ \\

\hline
$10 \epsilon$ & 
$(A + \epsilon)^3$ & 
$\sin(4BA + \epsilon)$ & 
$\sign(C) \sin(2CA + \epsilon)$ & 
$\sin(4DA + \epsilon)$ \\

\hline
$\sin(30 \epsilon)$ & 
$(A + \epsilon)^3$ & 
$10 \epsilon$ & 
$\sin(4CA + \epsilon)$ & 
$\sign(D) \sfrac{1}{|D| + 0.1} + \epsilon$ \\

\hline
$10 \epsilon$ & 
$\sin(A^3 - A + \epsilon)$ & 
$(B + \epsilon)^3$ & 
$\sin(C^3 - C + \epsilon)$ & 
$\sin(2A) + \epsilon$ \\

\hline
$\sin(30 \epsilon)$ & 
$\sign(A) \sfrac{1}{|A| + 0.1} + \epsilon$ & 
$\sign(B) \sfrac{1}{|B| + 0.1} + \epsilon$ & 
$\sin(2C) + \epsilon$ & 
$\sign(D) \sin(4DA + \epsilon)$ \\

\hline
$\sin(30 \epsilon)$ & 
$\sign(A) \sfrac{1}{|A| + 0.1} + \epsilon$ & 
$\sin(2B^3 -A^2) + \epsilon$ & 
$10 \epsilon$ & 
$\sign(D) \sfrac{1}{|D| + 0.1} + \epsilon$ \\

\hline
$10 \epsilon$ & 
$\sin(30 \epsilon)$ & 
$\sign(B) \sin(2BA + \epsilon)$ & 
$\sin(2C) \sin(\sfrac{1}{|B| + 0.1}) + \epsilon$ & 
$\sin(4DA + \epsilon)$ \\

\hline
$\sin(30 \epsilon)$ & 
$\sin(\sfrac{1}{|A| + 0.1} + \epsilon)$ & 
$(B + \epsilon)^3$ & 
$\sin(2B^3 -A^2) + \epsilon$ & 
$\sin(B^3 - B + \epsilon)$ \\

\hline
$\sin(30 \epsilon)$ & 
$\sin(\sfrac{1}{|A| + 0.1} + \epsilon)$ & 
$\sin(B^3 - B + \epsilon)$ & 
$\sin(\sfrac{1}{|B| + 0.1} + \epsilon)$ & 
$\sin(30 \epsilon)$ \\

\hline
\end{tabular}}
\vskip0.2cm
\caption{Generating equations used for the experiments presented in Table \ref{table:OurNonlinearity} ($\epsilon \sim \NN(0, 0.1^2)$ for each table cell independently).}
\label{table:nonlinear.generation}
\end{table}

\section{Relation between $\log \, p_\text{Bayes}(\DD\,|\,\GG)$ and $\log \, p_{\PI}(\DD\,|\,\GG)$}
\label{appsec:relation}

\begin{wrapfigure}[9]{l}{0.3\textwidth}
\centering
\vskip-0.5cm
\scalebox{0.9}{
\begin{tikzpicture}[dgraph]
\node at (-1,1) {$(a)$}; 
\node[dot] (Theta1) [label= north:$\theta^1$] at (0,2) {};
\node[dot] (Theta2) [label= north:$\theta^2$] at (1.5,2) {};
\node[dot] (Theta3) [label= north:$\theta^3$] at (3,2) {};
\node[dot] (x11) [label= south:$X^1$] at (0,1) {};
\node[dot] (x21) [label= south:$X^2$] at (1.5,1) {};
\node[dot] (x31) [label= south:$X^3$] at (3,1) {};
\draw[line width=0.6pt](x11)--(x21);
\draw[line width=0.6pt](x21)--(x31);
\draw[line width=0.6pt](Theta1)--(x11);
\draw[line width=0.6pt](Theta2)--(x21);
\draw[line width=0.6pt](Theta3)--(x31);
\end{tikzpicture}}
\scalebox{0.9}{
\begin{tikzpicture}[dgraph]
\node at (-1,1) {$(b)$}; 
\node[dot] (Theta1) [label= north:$\theta^1$] at (0,2) {};
\node[dot] (Theta2) [label= north:$\theta^2$] at (1.5,2) {};
\node[dot] (Theta3) [label= north:$\theta^3$] at (3,2) {};
\node[dot] (x11) [label= south:$X^1_1$] at (0,1) {};
\node[dot] (x21) [label= south:$X^2_1$] at (1.5,1) {};
\node[dot] (x31) [label= south:$X^3_1$] at (3,1) {};
\node[dot] (x12) [label= south:$X^1_2$] at (0,0) {};
\node[dot] (x22) [label= south:$X^2_2$] at (1.5,0) {};
\node[dot] (x32) [label= south:$X^3_2$] at (3,0) {};
\draw[line width=0.6pt](x11)--(x21);
\draw[line width=0.6pt](x21)--(x31);
\draw[line width=0.6pt](x12)--(x22);
\draw[line width=0.6pt](x22)--(x32);
\draw[line width=0.6pt](Theta1)--(x11);
\draw[line width=0.6pt](Theta2)--(x21);
\draw[line width=0.6pt](Theta3)--(x31);
\draw[->, line width=0.6pt](Theta1)to [bend right=+48] (x12);
\draw[->, line width=0.6pt](Theta2)to [bend right=+48] (x22);
\draw[->, line width=0.6pt](Theta3)to [bend right=+48] (x32);
\end{tikzpicture}}
\end{wrapfigure}

In this section we demonstrate that, if $p(\prm)=\prod_{d=1}^D p(\prm^d)$,
$\log \, p_\text{Bayes}(\DD \,|\,\GG)$ can be written as (omitting dependence on $\GG$)
\begin{align}
    \log \, p_\text{Bayes}(\DD)=\sum_{d=1}^D\log \,\prod_{i=1}^n 
    \int_{\Theta^d}  p(x^d_i  \,|\, \pa(x^d_i), \theta^d)  p(\theta^d\,|\,  \xli)\,  d\theta^d,
    \label{eq:factorization-bayes3}  
\end{align}
highlighting the connection with the factorization of $p_\text{plug-in}(\DD)$ over \cpds{} (\eqref{eq:PrequentialMDLDecomposed})
\begin{align*}
    \log \, p_{\PI}(\DD)&=\sum_{d=1}^D\log \,\prod_{i=1}^n p(x_i^d \,|\, \pa(x_i^d), \hat{\prm}^d(x_{<i})).
\end{align*}

To simplify the understanding but without loss of generality, we show the validity of \eqref{eq:factorization-bayes3} for the Bayesian network (a).
Let $\DD=\{x_i:=(x^1_i,x^2_i,x^3_i)\}_{i=1}^2$ be a dataset formed by two samples from $\int_{\Theta}p(X^1,X^2,X^3,\theta)d\theta$. We can view $\DD$ as  
formed by one sample from $\int_{\Theta}p(X_1,X_2,\theta)d\theta$, where $p(X_1,X_2,\theta)=p(X_1 \,|\, \theta)p(X_2 \,|\, \theta)p(\theta)$ with $p(X_1 \,|\, \theta)=p(X_2 \,|\, \theta)$, and $p(X_i:=(X^1_i,X^2_i,X^3_i)|\theta)=p(X^3_i \,|\, X^2_i,\theta^3)p(X^2_i \,|\, X^1_i,\theta^2)p(X^1_i \,|\, \theta^1)$, for $i=1,2$ (Bayesian network (b)). 
Using the prequential formulation $\log p(x_1,x_2)= \log \,p(x_2|x_1)p(x_1)$, $\log \, p_\text{Bayes}(\DD):=\log \int_{\Theta}p(x_1,x_2,\theta)d\theta$ can be written as 
\begin{align}
    \log \, p_\text{Bayes}(\DD) & %
    =\log \,p(x_2|x_1)p(x_1)%
    = \log \,\Bigl\{\int_{\Theta}p(x_2,\theta\,|\,x_1)d\theta\Bigr\}\Bigl\{\int_{\Theta}p(x_1,\theta)d\theta\Bigr\}\nonumber\\ 
    &= \log \,\Bigl\{\int_{\Theta}p(x_2\,|\,\theta,\cancel{x_1})p(\theta\,|\,x_1)d\theta\Bigr\}\Bigl\{\int_{\Theta}p(x_1\,|\,\theta)p(\theta)d\theta\Bigr\}\nonumber\\ 
    &= \log \,\Bigl\{\int_{\Theta}p(x^3_2\,|\,x^2_2,\theta^3)p(x^2_2\,|\,x^1_2,\theta^2)p(x^1_2\,|\,\theta^1)p(\theta^1\,|\,x_1)p(\theta^2\,|\,x_1)p(\theta^3\,|\,x_1)d\theta\Bigr\} \nonumber\\
    &\hskip0.5cm\times 
    \Bigl\{\int_{\Theta}p(x^3_1\,|\,x^2_1,\theta^3)p(x^2_1\,|\,x^1_1,\theta^2)p(x^1_1\,|\,\theta^1)p(\theta^1)p(\theta^2)p(\theta^3)d\theta\Bigr\}\nonumber\\      
    &=\sum_{d=1}^D\log \,\prod_{i=1}^n %
    \int_{\Theta^d}  p(x^d_i  \,|\, \pa(x^d_i), \theta^d)  p(\theta^d\,|\, \xli)\,  d\theta^d,
\label{eq:BayesFactorization2}    
\end{align}
where we have used the fact that $p(x_2\,|\,\theta,x_1)=p(x_2\,|\,\theta)$ and that the parameters posterior factorizes over $d$, i.e. $p(\theta\,|\,x_1)=p(\theta^1\,|\,x_1)p(\theta^2\,|\,x_1)p(\theta^3\,|\,x_1)$. 

In the case in which the distribution of the parameters for each \cpd~and value of parents is Dirichlet with parameter $\alpha$, 
a well-known result is
\begin{align*}
\int_{\Theta^d}  p(x^d_i =k \,|\, \pa(x^d_i)=l, \theta^d)  p(\theta^d\,|\, \xli)\,  d\theta^d = \frac{N_{k,l} + \alpha}{\sum_m (N_{m,l} + \alpha)}.
\end{align*}

\section{Prequential Scoring with Interventional Data}
\label{sec:tabularInterventions}

\begin{figure*}[t]
    \includegraphics[width=\textwidth]{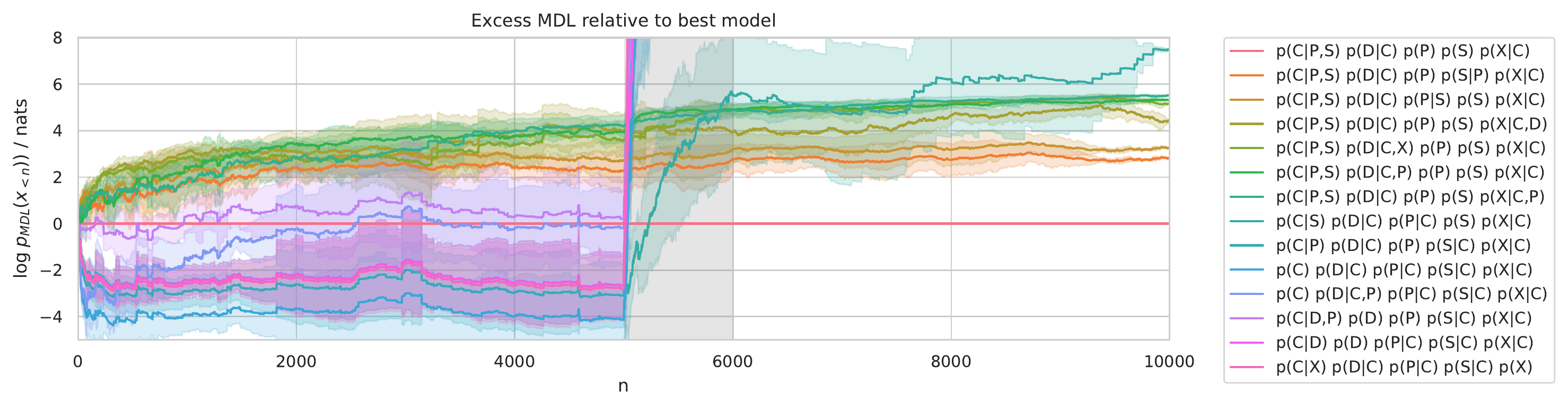}
    \vspace{-0.5cm}
    \caption{Excess log-loss in nats for $i=1,\ldots,10,000$ for the 14 DAGs with lowest excess.
    Uncertainty bands show standard deviation from 5 different different permutations of the data.
    Between $5,000 \le i < 6,000$ (shaded area) interventional samples were supplied:   
    we replaced the value of a random node in the graph with an random value (p = \sfrac{1}{2}).
    Only due to these interventional samples could the ground-truth DAG $\GG^*$ be correctly identified as most likely.
    Nevertheless,  there is no strong evidence in favour of a structure as many of them differ by only a few nats.}
    \label{fig:cancer}
\end{figure*}

In \figref{fig:cancer} we show the influence of interventional data on \method{}. We generated observations
from the cancer Bayesian network from the bnlearn repository \citep{erdogant2019bnlearn} with 
DAG $\GG^*=P\rightarrow C \leftarrow S, C \rightarrow X, C \rightarrow D$. 
For $i\in\{5,000,\ldots, 6000\}$, we generated interventional data by randomly selecting a node from the graph and replacing it with a random value. The excess prequential loss in nats is plotted for the 14 best ranking DAGs during the first half of the sequence and the 14 best after learning from all observations. The interventional samples have a strong influence on the ranking and are necessary for the correct identification of $\GG^*$.

\end{document}